\documentclass{article}
\usepackage{spconf,amsmath,graphicx}
\usepackage{float}
\usepackage{booktabs}
\usepackage{bigstrut}
\usepackage{multirow}
\usepackage{array}
\usepackage{amsthm,amssymb}
\usepackage{mathrsfs}
\usepackage{threeparttable}
\usepackage{multicol}
\usepackage{makecell}
\usepackage[colorlinks=true]{hyperref}

\usepackage{enumitem}
\setlist{nosep, leftmargin=14pt}

\usepackage{mwe} 

\title{A Foundation Model for General Moving Object Segmentation in Medical Images}
\name{%
\begin{tabular}{@{}c@{}}
Zhongnuo Yan$^{1,2,3\star}$ \qquad 
Tong Han$^{1,2,3\star}$ \qquad 
Yuhao Huang$^{1,2,3}$ \qquad 
Lian Liu$^{1,2,3}$ \qquad 
Han Zhou$^{1,2,3}$ \\
Jiongquan Chen$^{1,2,3}$ \qquad 
Wenlong Shi$^4$ \qquad 
Yan Cao$^4$ \qquad 
Xin Yang$^{1,2,3\dagger}$ \qquad 
Dong Ni$^{1,2,3\dagger}$ \qquad 
\end{tabular}\thanks{$^\star$Contribute equally to this work. \hspace*{2em}$^\dagger$Corresponding author: xinyang@szu.edu.cn (Xin Yang), nidong@szu.edu.cn (Dong Ni).}}

\address{$^1$ National-Regional Key Technology Engineering Laboratory for Medical Ultrasound,\\ School of Biomedical Engineering, Shenzhen University Medical School, Shenzhen University, China \\
$^2$Medical Ultrasound Image Computing (MUSIC) Lab, Shenzhen University, China \\
$^3$Marshall Laboratory of Biomedical Engineering, Shenzhen University, China \\
$^4$Shenzhen RayShape Medical Technology Inc., Shenzhen, China}

\begin{document}
%
\maketitle
\begin{abstract}
Medical image segmentation aims to delineate the anatomical or pathological structures of interest, playing a crucial role in clinical diagnosis. 
A substantial amount of high-quality annotated data is crucial for constructing high-precision deep segmentation models.
However, medical annotation is highly cumbersome and time-consuming, especially for medical videos or 3D volumes, due to the huge labeling space and poor inter-frame consistency.
Recently, a fundamental task named Moving Object Segmentation (MOS) has made significant advancements in natural images. 
Its objective is to delineate moving objects from the background within image sequences, requiring only minimal annotations. 
In this paper, we propose the first foundation model, named iMOS, for MOS in medical images.
Extensive experiments on a large multi-modal medical dataset validate the effectiveness of the proposed iMOS.
Specifically, with the annotation of only one image in the sequence, iMOS can achieve satisfactory tracking and segmentation performance of moving objects throughout the entire sequence in bi-directions.
We hope that iMOS can help accelerate the annotation speed of experts, and boost the development of medical foundation models.  


\end{abstract}

\section{Introduction}
\label{sec:intro}

\begin{figure}[t]
\centering
\includegraphics[width=1.0\linewidth]{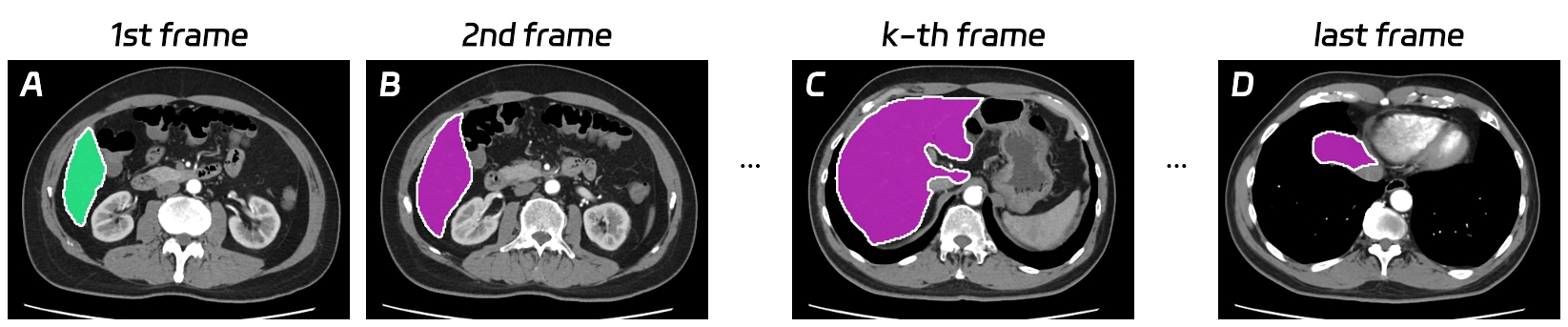}
\caption{Process of semi-supervised MOS. 
\textbf{(A)} represents the manual annotation. \textbf{(B)-(D)} are segmentation results.} 
\vspace{-0.4cm}
\label{fig:pipeline}
\end{figure}

The accurate segmentation of specific regions in medical images is crucial for clinical diagnosis and treatment planning. 
Recently, deep learning-based segmentation algorithms have shown promising results in medical images~\cite{ronneberger2015u}. 
Most of them focused on the segmentation of only one single modality or specific anatomical structure.
Different from them, the Segment Anything Model (SAM)~\cite{kirillov2023segment} was proposed to achieve general image segmentation, and lots of its extension works analyzed the medical SAM~\cite{ma2024segment,huang2024segment}. 
In their methods, using a few manual prompts 
can guide the SAM to achieve impressive segmentation accuracy on specific medical images.
However, directly applying these 2D models to videos or 3D volumes often presents additional challenges~\cite{yang2023track}.
Moreover, they required substantial manual prompts for segmenting multiple frames/slices or 3D volumes.
This limits their flexibility and convenience in real-life usage scenarios.

Most recently, several studies have been proposed to explore the novel task, named Moving Object Segmentation (MOS).
It aims to automatically track and segment moving objects in consecutive frames with minimal or no annotations (i.e., semi-supervised and unsupervised MOS).
In this study, we focus on exploring the semi-supervised MOS (see Figure~\ref{fig:pipeline}), due to the structure complexity and scene variability of medical data.
Compared to the unsupervised MOS, the semi-supervised one can ensure the accuracy of segmentation while maintaining flexibility. 
Semi-supervised MOS has already demonstrated impressive performance in natural images~\cite{seoung2019video,cheng2022xmem}. 
However, there is still a lack of research that investigates MOS in medical images currently.

In this work, we propose a foundation model for general MOS in medical images named iMOS.
To the best of our knowledge, we are the first to explore MOS in medical images. Additionally, we validated the proposed method on a multi-modal and large-scale medical dataset. 
By labeling one single frame, iMOS can achieve forward and backward segmentation in the sequence, even for those objects unseen during training.
This proves the strong generalization of our proposed foundation MOS model. Code have been released at~\url{https://github.com/yzhn16/iMOS}.

\section{Related Work}
\label{sec:related work}

Current MOS can primarily be categorized into unsupervised and semi-supervised approaches. 
Different from general deep learning, the supervision here only refers to manual interaction during testing rather than the training~\cite{perazzi2016benchmark}.

\noindent{\textbf{Unsupervised MOS.}} 
It aims to segment salient object sequences without any human intervention.
Traditional methods used similarity calculation \cite{faktor2014video}, motion boundaries \cite{ochs2013segmentation}, and thresholds \cite{zhu2008novel}, etc., to segment objects. 
These methods were easily affected by occlusion, contrast, and complex structures. 
Deep-learning-based methods can better model object motion information using techniques including optical flow~\cite{zhou2020motion}, temporal coherence~\cite{yang2019anchor}, and motion information enhancement~\cite{ji2021full}. 
However, due to the lack of supervision, the above methods were poor in robustness and effectiveness.

\noindent{\textbf{Semi-supervised MOS.} }
In this kind of method, the masks of the first frame or the first few frames were first given for supervision.
Then, the model segmented subsequent frames based on the cues of these frames. 
Representative approaches relied on data augmentation~\cite{caelles2017one}, yet their effectiveness was limited, and they often suffered from slow processing speeds.
Tracking-based methods~\cite{xiao2018monet} were proficient at monitoring moving objects. However, they tended to accumulate errors over time, resulting in diminished robustness.
At present, memory-based methods~\cite{seoung2019video,cheng2022xmem} have emerged as the primary approach for MOS, primarily owing to their outstanding performance and straightforward architecture.
Nevertheless, the above methods were all constructed for natural image segmentation. 
Medical images present unique characteristics and challenges compared to natural images. Due to the critical need for achieving high accuracy and efficiency in medical image segmentation,
it is highly desired for the establishment of a foundational medical MOS model.

\section{Methodology}
\label{sec:method}

\subsection{Semi-supervised MOS}
\begin{figure}[!t]
\centering
\includegraphics[width=1.0\linewidth]{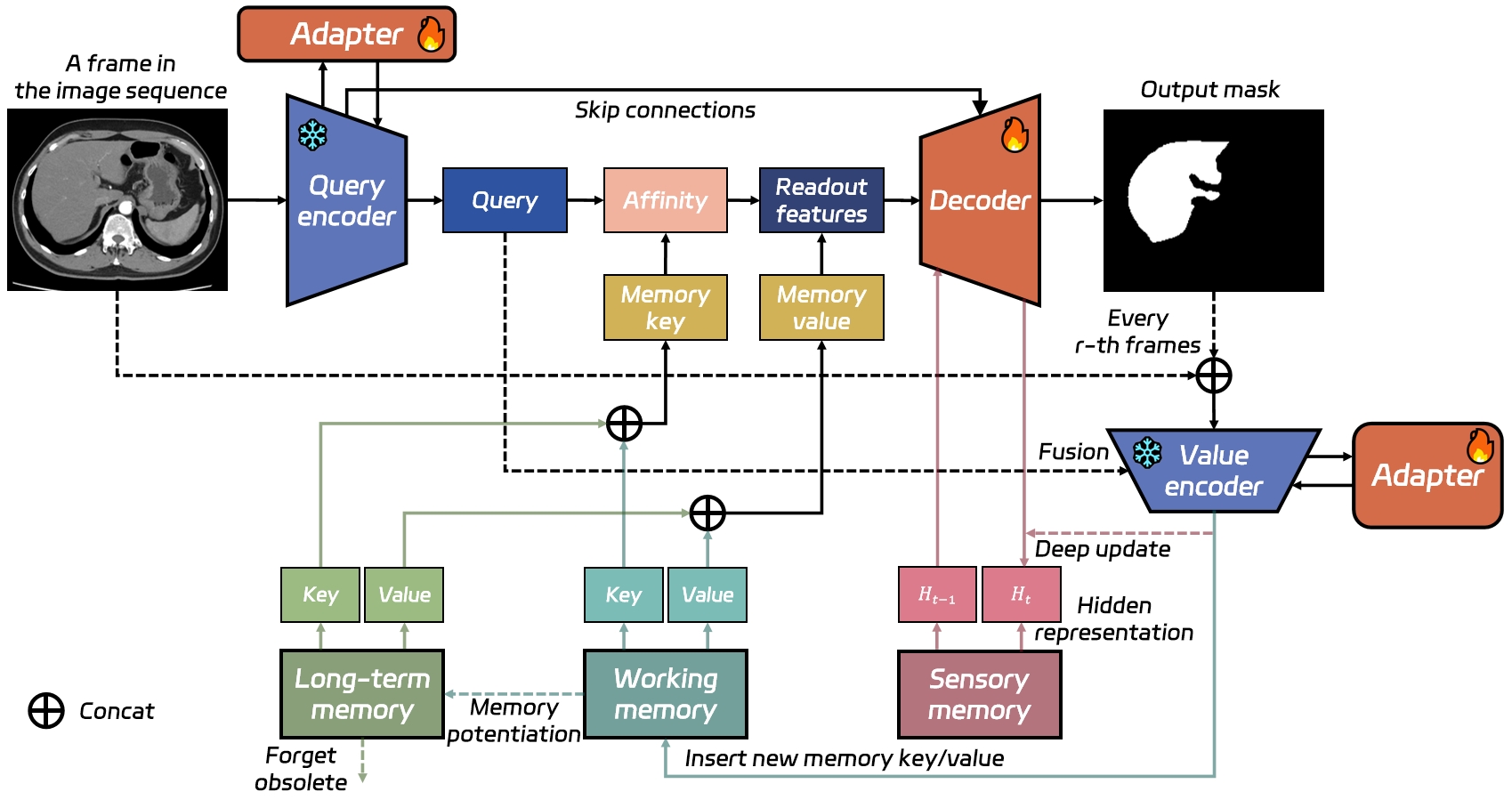}
\caption{Architecture of iMOS. Lines with different colors are pathways of different memory systems. } \label{fig:arch-main}
\end{figure}

We consider the segmentation of anatomical structures in medical images, whether in videos or 3D volumes, as a semi-supervised MOS task. We consider frames in video or slices in 3D image which contain the anatomical structures of interest as a image sequence. Additionally, manual annotations for the first frame are provided, and the segmentation of subsequent images is completed by MOS method. Figure~\ref{fig:arch-main} illustrates the architecture of our method which is based on XMem~\cite{cheng2022xmem}, it only requires manual annotation of a single frame to complete the segmentation for all frames. 

\subsubsection{Memory Reading}

iMOS consists of three trainable modules: query encoder, decoder, value encoder, and three memory systems: sensory memory, working memory, long-term memory. Given a sequence of images and the mask of the target objects in the first frame, sensory memory and working memory are initialized with the first frame and its corresponding mask. For subsequent images, readout features $F$ are extracted from the memory value $v$ based on the affinity $W$ between query $q$ and memory key $k$. Simultaneously, a mask is generated by the decoder with $F$ and the hidden representation, $h_{t-1}$, from the previous time step in the sensory memory. Additionally, the value encoder encodes the image and mask every $r$-th frames to generate new memory value. The memory reading process can be expressed as: $F=vW(k,q)$.

\subsubsection{Memory Updating}
The memory systems in iMOS update at various frequencies to store features at different time scales. Sensory memory updates at the fastest frequency to retain short-term information, utilizing multi-scale features from the decoder at every frame and updating through gated recurrent units (GRU). Furthermore, at every $r$-th frame, the memory value generated by the value encoder are used to update sensory memory via another GRU, enabling sensory memory to discard redundant information. Working memory is employed for precise matching over a few seconds, where memory value and memory key copied from query are added to the working memory every $r$-th frames. To avoid excessive memory use, the working memory frame count is limited between $T_{min}$ and $T_{max}$. Long-term memory stores compact and representative features. When working memory reaches its storage limit at $T_{max}$, all frames except the first and the most recent $T_{min}-1$ frames are added to long-term memory via a prototype selection and memory potentiation algorithm.

\noindent{\textbf{Prototype Selection.}~Prototype selection aims to choose a representative subset, denoted as prototype keys $k^p$, from the candidate keys $k^c$. These subsets have the highest usage frequency within working memory, which is defined by the cumulative total affinity in the affinity matrix $W$.}

\noindent{\textbf{Memory Potentiation.}~The prototype selection results in sparse prototype keys. If the same method were applied to extract prototype values, it might not adequately represent other candidates. Memory potentiation is introduced to address this issue. Specifically, similar to memory reading, candidate value are extracted with affinity matrix $W(k^c,k^p)$ between candidate key $k^c$ and prototype key $k^p$: $v^p=v^cW(k^c,k^p)$}
\subsection{Parameter Efficient Tuning}

\begin{figure}[htbp]
\centering
\includegraphics[width=1.0\linewidth]{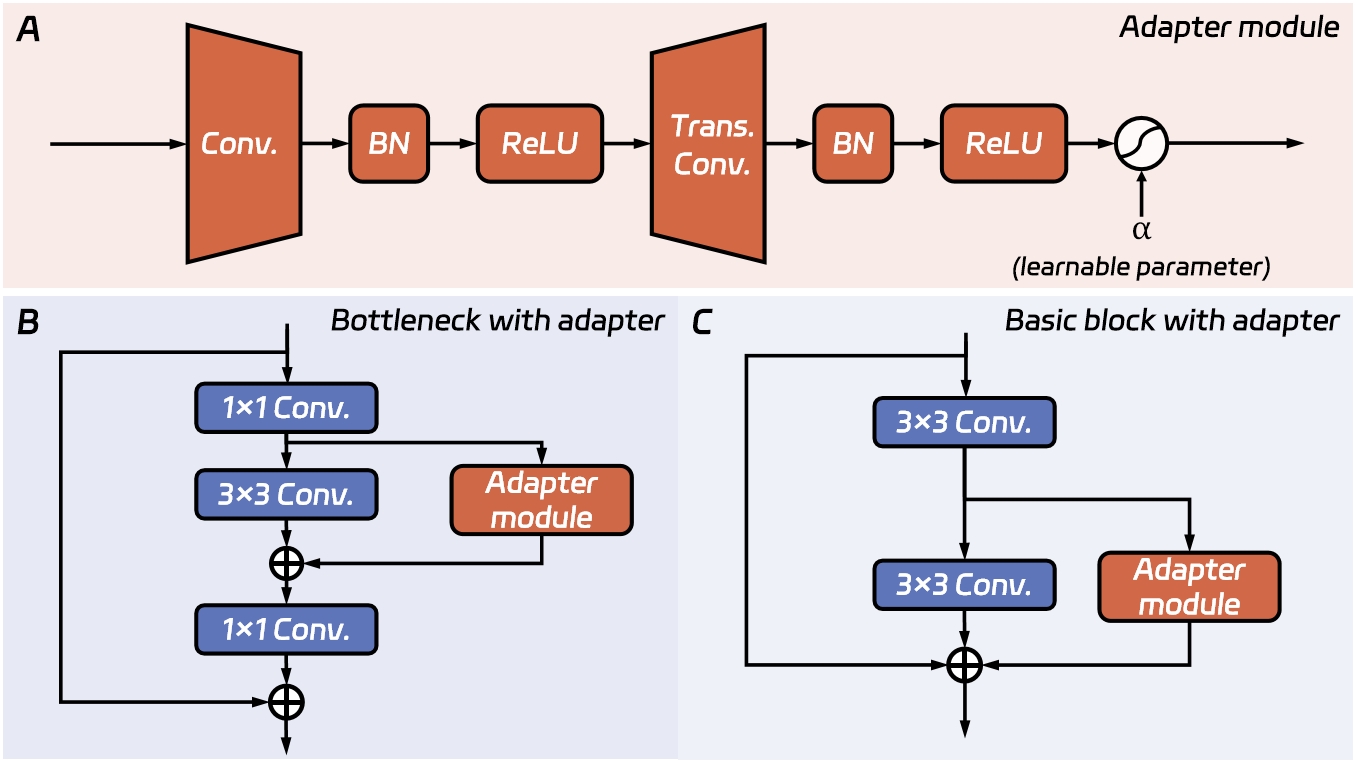}
\vspace{-0.5cm}
\caption{Architecture (\textbf{A}) and implementation scheme (\textbf{B}, \textbf{C}) of adapter module. $\alpha$ controls output weight; (\textbf{B}) and (\textbf{C}) are residual blocks in query and value encoder, respectively.} \label{fig:adapter-arch-scheme}
\vspace{-0.3cm}
\end{figure}

The training process of iMOS includes three phases: training on static images, training on long sequences, and training on short sequences. Training the model from scratch or fine-tuning the entire model is time-consuming. To facilitate rapid and efficient fine-tuning of the model on large-scale dataset, we introduced the adapter module based on the Conv. Parallel structure mentioned in~\cite{chen2022convadapter}, which achieves the best trade-off between parameter efficiency and performance. We kept the query encoder and value encoder parameters fixed and then applied the adapter to them, enabling them to learn residual representations between medical images and natural images. The architecture and implementation scheme of adapter module are illustrated in Figure~\ref{fig:adapter-arch-scheme}.

\section{Experimental Results}
\label{sec:exp}

\subsection{Dataset and Training Details}
\begin{figure}[!t]
\centering
\includegraphics[width=0.95\linewidth]{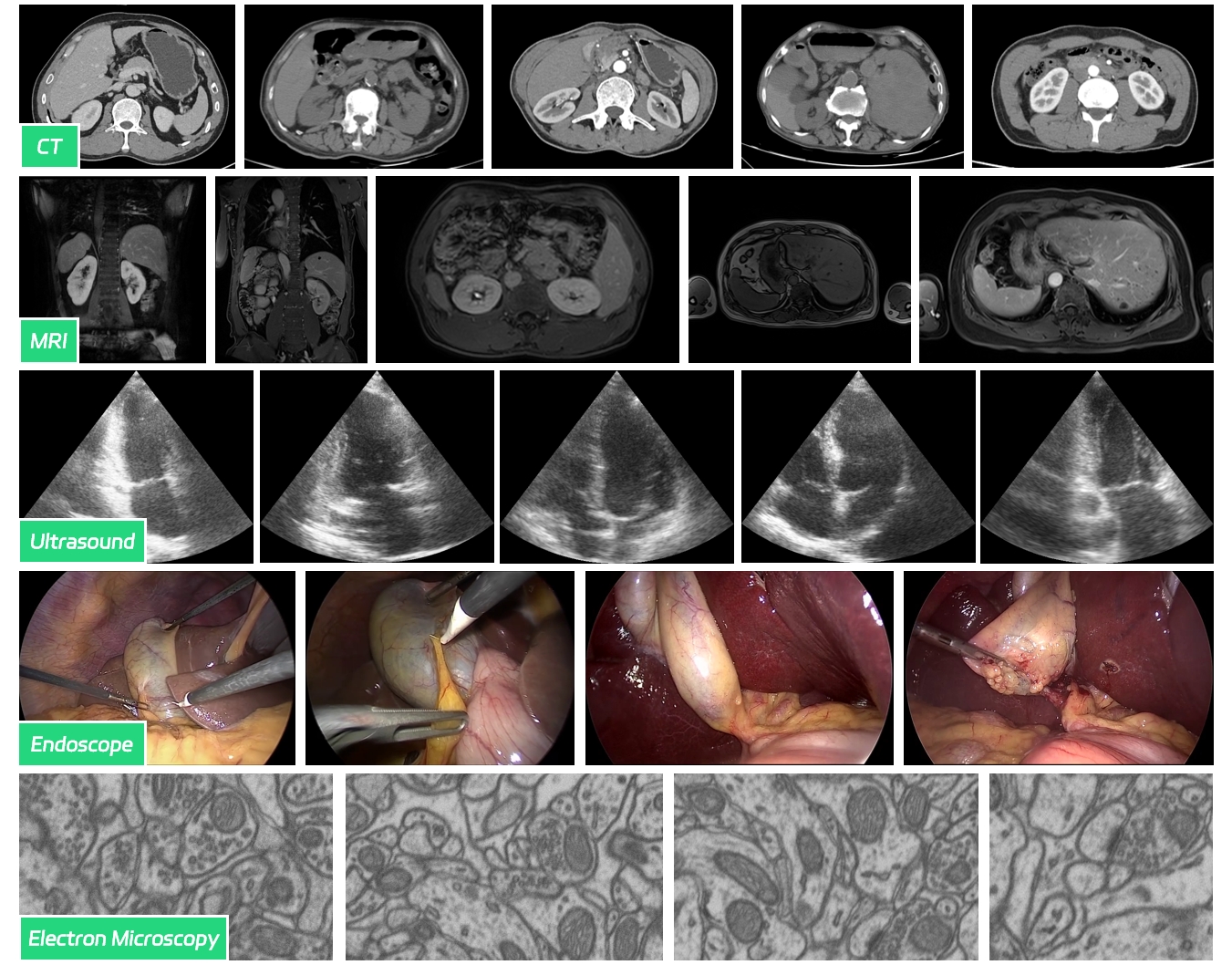}
\vspace{-0.5cm} 
\caption{Examples of the five modalities in the dataset.} \label{fig:eg}
\end{figure}

To evaluate the performance of iMOS on medical images, we collected data from multiple public datasets~\cite{ji2022amos,leclerc2019deep,hong2020cholecseg8k,lucchi2013learning}, including five modalities, i.e., CT, MRI, ultrasound, endoscopy, and electron microscopy (see Figure~\ref{fig:eg}). For each category of video and 3D volume, we constructed a dataset in which objects are tracked from the first frame to the last frame. 
Details of the dataset are shown in Table~\ref{tab:dataset}. 

\begin{table}[!t]
\vspace{-0.5cm} 
\caption{Dataset description.}
\label{tab:dataset}
\resizebox{\linewidth}{!}{
\begin{tabular}{ccccc}
\Xhline{2px}
\textbf{Modality}   & \textbf{Source}         & \textbf{Category} & \textbf{Video/3D volume} & \multicolumn{1}{l}{\textbf{Generated sequence}} \\ \hline
CT & \multirow{2}{*}{AMOS22~\cite{ji2022amos}} & 7 & 300 & 2091  \\
MRI & & 7 & 60 & 419  \\
Ultrasound & CAMUS~\cite{leclerc2019deep} & 3 & 1000 & 3000  \\
Endoscopy & CholecSeg8k~\cite{hong2020cholecseg8k} & 12 & 101 & 597 \\
Electron Microscopy & EPFL~\cite{lucchi2013learning} & 1 & 2 & 109 \\
\Xhline{2px}
\vspace{-0.7cm}
\end{tabular}
}
\end{table}

 \begin{table}[!b]
  \centering
  \vspace{-0.7cm} 
  \caption{Validity experiment.}

  \resizebox{\linewidth}{!}{
    \begin{tabular}{c|ccc|ccc}
    \Xhline{2px}
    \multirow{2}[4]{*}{\textbf{Modality}} & \multicolumn{3}{c|}{\textbf{Before Fine-tuning}} & \multicolumn{3}{c}{\textbf{After Fine-tuning}} \bigstrut\\
\cline{2-7}          & $\mathcal{J}$\&$\mathcal{F}$↑  & $\mathcal{J}$↑     & $\mathcal{F}$↑     & $\mathcal{J}$\&$\mathcal{F}$↑  &  $\mathcal{J}$↑    & $\mathcal{F}$↑ \bigstrut\\
    \hline
    CT   & 0.716(0.274)  & 0.627(0.307)   &  0.804(0.255)   & 0.850(0.201)   &  0.800(0.227)   & 0.904(0.188) \bigstrut[t]\\
    MRI  &  0.626(0.292) &  0.542(0.311)  & 0.710(0.291)   &   0.806(0.225)   &  0.759(0.256)    &  0.854(0.208) \\
    Ultrasound & 0.702(0.168)  & 0.777(0.141)  &  0.628(0.213)   & 0.859(0.120)  &  0.882(0.071)    & 0.836(0.180) \\
    Endoscope & 0.850(0.162)  & 0.853(0.142)  &  0.852(0.140)   &     0.856(0.138)  &  0.854(0.160)    & 0.858(0.140) \\
    Electron Microscopy   &   0.732(0.222) &  0.619(0.247)     &  0.846(0.206)     &  0.879(0.144)    &  0.805(0.149)     & 0.954(0.143) \bigstrut[b]\\
    \Xhline{2px}
    \end{tabular}%
    }
      \label{Table:Validity}%
\end{table}%

\begin{table*}[!t]
\caption{Generalizability experiment. Categories marked with asterisk ($^*$) indicate that they did not appear in the training set.}
\label{tab:generalizability}
\resizebox{\textwidth}{!}{
\begin{tabular}{ccccc||ccccc}
\Xhline{2px}
\textbf{Modality} & \textbf{Category} & $\mathcal{J}\&\mathcal{F}\uparrow$ & $\mathcal{J}\uparrow$ & $\mathcal{F}\uparrow$ & \textbf{Modality} & \textbf{Category} & $\mathcal{J}\&\mathcal{F}\uparrow$ & $\mathcal{J}\uparrow$ & $\mathcal{F}\uparrow$ \\ \hline
\multicolumn{1}{c|}{\multirow{7}{*}{CT}} & Spleen & 0.893(0.168) & 0.855(0.196) & 0.932(0.160) & \multicolumn{1}{c|}{\multirow{3}{*}{Ultrasound}} & Left atrium & 0.888(0.119) & 0.903(0.068) & 0.872(0.176) \\
\multicolumn{1}{c|}{} & Right kidney & 0.909(0.133) & 0.871(0.165) & 0.948(0.126) & \multicolumn{1}{c|}{} & Left ventricle epicardium & 0.885(0.094) & 0.874(0.069) & 0.897(0.126) \\
\multicolumn{1}{c|}{} & Left kidney & 0.904(0.158) & 0.862(0.191) & 0.945(0.134) & \multicolumn{1}{c|}{} & Left ventricle endocardium* & 0.896(0.104) & 0.924(0.048) & 0.869(0.165) \\ \cline{6-10} 
\multicolumn{1}{c|}{} & Esophagus* & 0.810(0.200) & 0.683(0.214) & 0.937(0.143) & \multicolumn{1}{c|}{\multirow{12}{*}{Endoscope}} & Abdominal wall & 0.886(0.057) & 0.928(0.085) & 0.844(0.110) \\
\multicolumn{1}{c|}{} & Liver* & 0.782(0.270) & 0.760(0.395) & 0.804(0.258) & \multicolumn{1}{c|}{} & Liver & 0.895(0.054) & 0.910(0.087) & 0.880(0.103) \\
\multicolumn{1}{c|}{} & Stomach* & 0.780(0.208) & 0.752(0.243) & 0.809(0.231) & \multicolumn{1}{c|}{} & Gastrointestinal tract & 0.816(0.126) & 0.811(0.170) & 0.822(0.153) \\
\multicolumn{1}{c|}{} & Arota* & 0.863(0.169) & 0.800(0.227) & 0.945(0.140) & \multicolumn{1}{c|}{} & Fat & 0.872(0.062) & 0.906(0.095) & 0.837(0.097) \\ \cline{1-5}
\multicolumn{1}{c|}{\multirow{7}{*}{MRI}} & Spleen & 0.921(0.091) & 0.885(0.126) & 0.956(0.091) & \multicolumn{1}{c|}{} & Grasper & 0.818(0.162) & 0.750(0.198) & 0.887(0.149) \\
\multicolumn{1}{c|}{} & Right kidney & 0.917(0.078) & 0.893(0.104) & 0.940(0.083) & \multicolumn{1}{c|}{} & Gallbladder & 0.876(0.161) & 0.875(0.142) & 0.861(0.171) \\
\multicolumn{1}{c|}{} & Left kidney & 0.907(0.100) & 0.877(0.131) & 0.937(0.106) & \multicolumn{1}{c|}{} & Connective tissue* & 0.837(0.065) & 0.858(0.095) & 0.816(0.146) \\
\multicolumn{1}{c|}{} & Esophagus* & 0.686(0.245) & 0.513(0.249) & 0.859(0.209) & \multicolumn{1}{c|}{} & Blood* & 0.636(0.205) & 0.583(0.220) & 0.689(0.182) \\
\multicolumn{1}{c|}{} & Liver* & 0.775(0.215) & 0.756(0.263) & 0.795(0.219) & \multicolumn{1}{c|}{} & Cystic duct* & 0.870(0.040) & 0.806(0.065) & 0.935(0.068) \\
\multicolumn{1}{c|}{} & Stomach* & 0.727(0.240) & 0.662(0.263) & 0.793(0.216) & \multicolumn{1}{c|}{} & L-hook electrocautery* & 0.912(0.080) & 0.906(0.099) & 0.918(0.106) \\
\multicolumn{1}{c|}{} & Arota* & 0.725(0.289) & 0.649(0.311) & 0.802(0.281) & \multicolumn{1}{c|}{} & Hepatic vein* & 0.729(0.186) & 0.556(0.207) & 0.901(0.098) \\ \cline{1-5}
\multicolumn{1}{c|}{Electron Microscopy} & Mitochondria & 0.879(0.144) & 0.805(0.149) & 0.954(0.143) & \multicolumn{1}{c|}{} & Liver ligament* & 0.945(0.004) & 0.973(0.007) & 0.918(0.043) \\ 
\Xhline{2px}
\end{tabular}}
\vspace{-0.4cm} 
\end{table*}

We split all of the video or 3D volume into a 3:1:1 ratio for training/validation/testing. All images were resized to a resolution of $480\times480$. Additionally, given the wide CT value range, we adjust appropriate window widths and window levels according to different anatomical regions. All other experimental settings remained consistent with those in XMem~\cite{cheng2022xmem}.

\begin{figure}[t]
\centering
\includegraphics[width=1.0\linewidth]{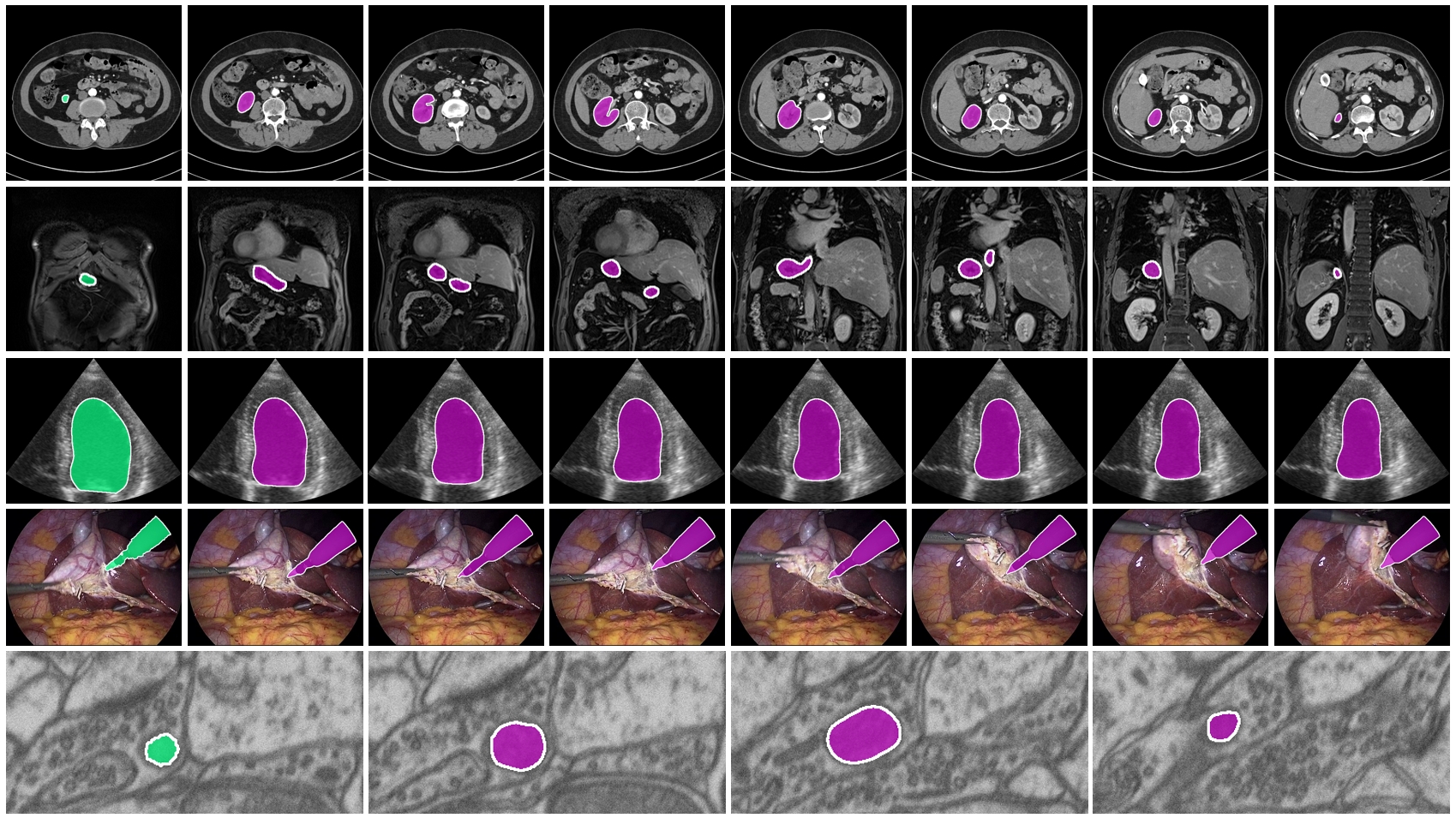}
\vspace{-0.4cm} 
\caption{Examples of segmentation results. The first image in the left column is the manually annotated mask.} 
\label{fig:good-case-all-modal}
\end{figure}

\begin{figure}[h]
\centering
\includegraphics[width=\linewidth]{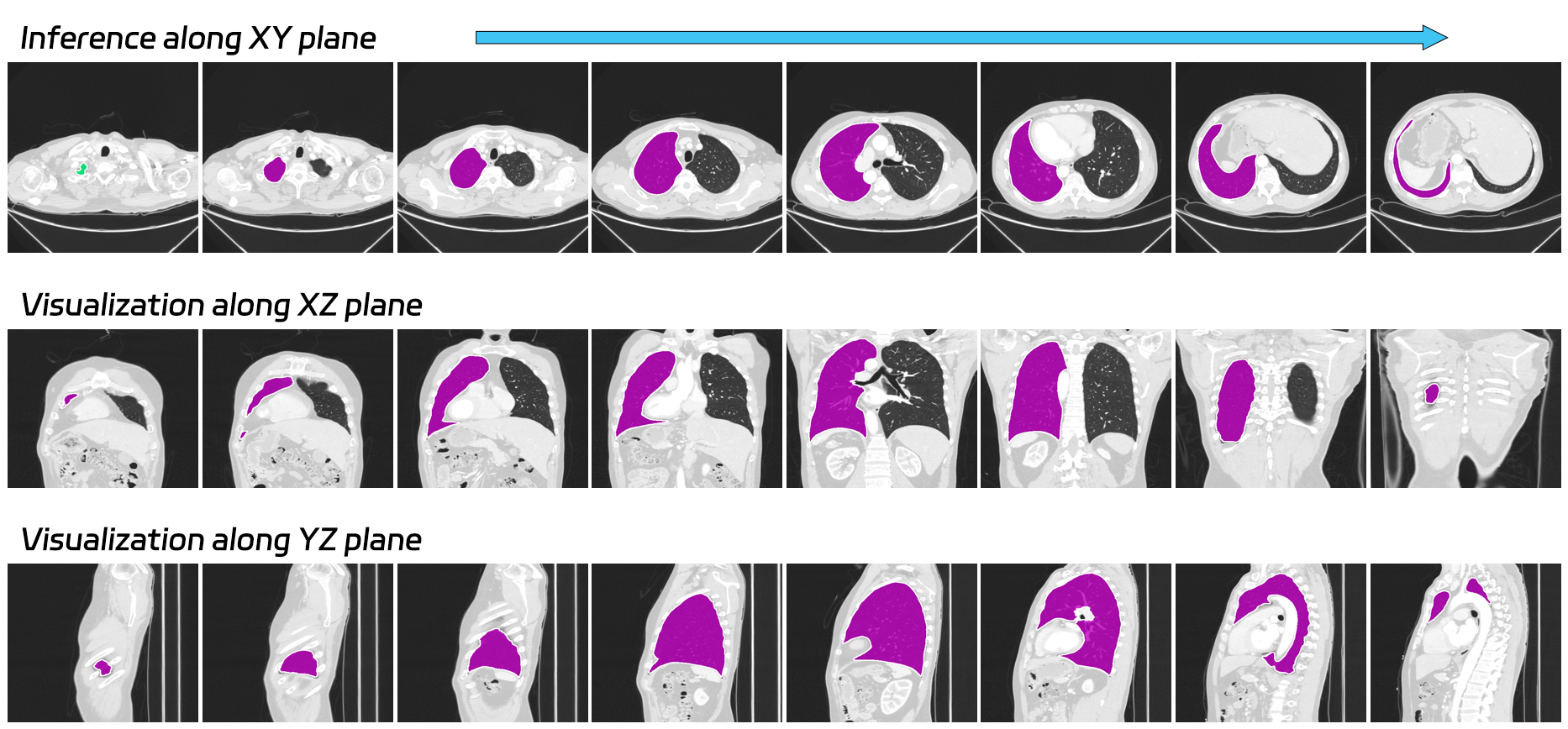}
\vspace{-0.5cm} 
\caption{Qualitative results of 3D segmentation of lung organs.  The results are based on the first
frame of transverse plane.}
\vspace{-0.3cm} 
\label{fig:lung-3axis}
\end{figure}

\begin{figure}[h]
\centering
\includegraphics[width=\linewidth]{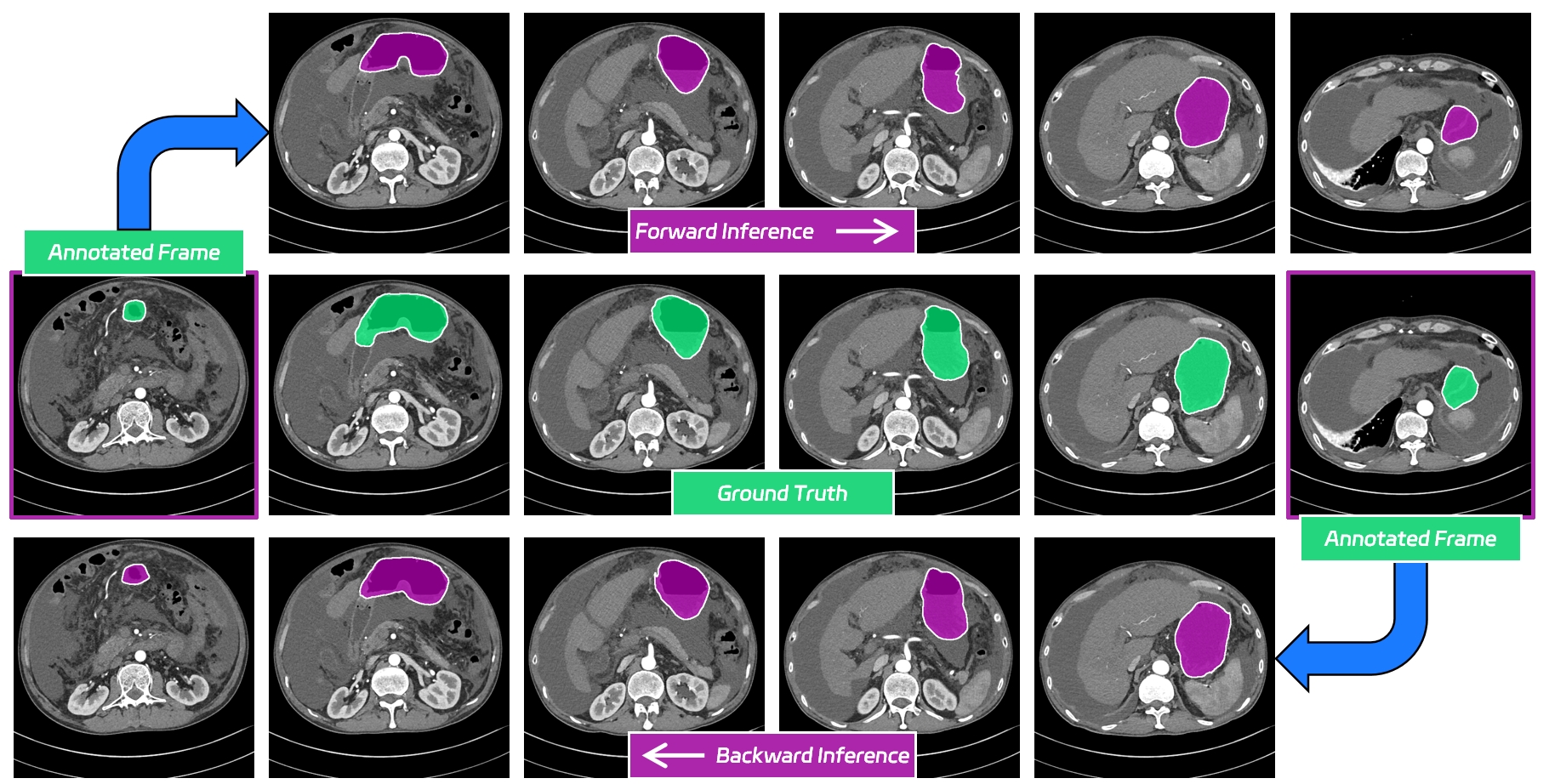}
\vspace{-0.5cm} 
\caption{Bi-directional segmentation results of stomach.} 
\vspace{-0.5cm} 
\label{fig:forward-backward}
\end{figure}

\subsection{Quantitative and Qualitative Evaluation}

We designed two experiments to test iMOS in terms of validity and generalizability, respectively.
First, we compared the performance of the model before and after fine-tuning on medical image segmentation. 
Second, we used the fine-tuned model to segment multiple organs in multiple modalities, including organs unseen during training.
Regional similarity $\mathcal{J}$ \cite{everingham2010pascal}, contour similarity $\mathcal{F}$ \cite{martin2004learning}, and their averages $\mathcal{J}$\&$\mathcal{F}$ were used as evaluation metrics for comprehensive assessment. 

Tables \ref{Table:Validity}  and \ref{tab:generalizability} illustrate the validity and generalizability of iMOS. 
In Table \ref{Table:Validity}, the performance of objects in all five modalities improves after model fine-tuning. 
Notably, the highest improvement can be observed in MRI, with an increase of 28.7\% on $\mathcal{J}\&\mathcal{F}$. Additionally, see Table~\ref{tab:generalizability}, almost all categories achieve $\mathcal{F}>$0.8, with some reaching 0.9, indicating that iMOS performs well in terms of contour accuracy.
For categories not present in the training set (marked with asterisk), iMOS still achieved good segmentation results, indicating its strong generalizability.
The experimental results confirm the effectiveness of the fine-tuning strategy and the generalization capability of iMOS across various categories.


Figure~\ref{fig:good-case-all-modal} illustrates the segmentation results on various modalities. With manual annotation of the first frame, iMOS can accurately track and segment objects in the image sequence. It performs well even with small or split objects (see the first two rows in Figure~\ref{fig:good-case-all-modal}). 

Figure~\ref{fig:lung-3axis} further illustrates the segmentation results of iMOS on 3D volume from different perspectives. Based on the initial slice of the 3D volume transverse plane (view the first image in the upper left corner), iMOS has the ability to segment all the slices of the 3D volume.
Even better, the iMOS segmentation outcomes exhibit favorable results when examined from both of the other planes.

Figure~\ref{fig:forward-backward} presents segmentation results for the same sequence in two different directions. 
In addition to segmentation based on the first frame mask, iMOS can reverse the segmentation of targets in previous frames through the mask of the last frame.
The segmentation results in the forward and backward directions are comparable.

Results above demonstrate iMOS's robust segmentation capabilities in both video and 3D volume, holding the potential to accelerate the annotation speed and optimize the annotation quality for experts.


\section{Conclusion}

In this work, we propose iMOS, a foundation model for general moving object segmentation in medical images. iMOS can automatically segment the remaining targets based on the manually-annotated mask of the first frame/slice in video/volume. 
We have demonstrated its validity and generalizability through extensive experiments. 
We believe that iMOS can provide technical accumulation for auxiliary medical video or 3D volume annotation tasks, and boost the development of the medical image analysis community.


\section{Compliance with ethical standards}
\vspace{-0.2cm}
The research was conducted using medical images in open access. Ethical approval was not required as confirmed by the license attached with the open access data.

\vspace{-0.3cm}
\section{Acknowledgements}
\vspace{-0.2cm}
This work was supported by the grant from National Natural Science Foundation of China (Nos. 62101343, 62171290), Shenzhen-Hong Kong Joint Research Program (No. SGDX20\newline201103095613036), Science and Technology Planning Project of Guangdong Province, China (2023A0505020002).

\vspace{-0.3cm}

\bibliographystyle{IEEEbib}
\bibliography{refs}
\end{document}